\documentclass[conference]{IEEEtran}
\ifCLASSINFOpdf
\else
\fi

\usepackage{cite}
\usepackage{graphicx}
\usepackage{latexsym}
\usepackage{amssymb}
\usepackage{url}
\usepackage[fleqn]{amsmath}
\usepackage[varg]{txfonts}
\usepackage{algpseudocode}
\usepackage{refcount}
\usepackage{footnote}
\usepackage{breqn}
\usepackage{hhline}
\usepackage{multirow}
\usepackage[utf8]{inputenc}
\usepackage[english]{babel}
\usepackage[table, x11names]{xcolor}

\usepackage{amsthm}

\theoremstyle{definition}
\newtheorem{definition}{Definition}[section]

\theoremstyle{remark}

\DeclareMathAlphabet\mathbfcal{OMS}{cmsy}{b}{n}

\begin{document}
%
% paper title
% Titles are generally capitalized except for words such as a, an, and, as,
% at, but, by, for, in, nor, of, on, or, the, to and up, which are usually
% not capitalized unless they are the first or last word of the title.
% Linebreaks \\ can be used within to get better formatting as desired.
% Do not put math or special symbols in the title.
\title{Tensor Decomposition for Compressing Recurrent Neural Network}

% author names and affiliations
% use a multiple column layout for up to three different
% affiliations
\author{\IEEEauthorblockN{Andros Tjandra\IEEEauthorrefmark{1}\IEEEauthorrefmark{2}, Sakriani Sakti\IEEEauthorrefmark{1}\IEEEauthorrefmark{2}, Satoshi Nakamura\IEEEauthorrefmark{1}\IEEEauthorrefmark{2}}
\IEEEauthorblockA{\IEEEauthorrefmark{1}Graduate School of Information Science, Nara Institute of Science and Techonology, Japan\\
\IEEEauthorrefmark{2} RIKEN, Center for Advanced Intelligence Project AIP, Japan\\
Email: \{andros.tjandra.ai6, ssakti, s-nakamura\}@is.naist.jp}
}

% conference papers do not typically use \thanks and this command
% is locked out in conference mode. If really needed, such as for
% the acknowledgment of grants, issue a \IEEEoverridecommandlockouts
% after \documentclass

% use for special paper notices
%\IEEEspecialpapernotice{(Invited Paper)}

% make the title area
\maketitle

% As a general rule, do not put math, special symbols or citations
% in the abstract
\begin{abstract}
In the machine learning fields, Recurrent Neural Network (RNN) has become a popular architecture for sequential data modeling. However, behind the impressive performance, RNNs require a large number of parameters for both training and inference. In this paper, we are trying to reduce the number of parameters and maintain the expressive power from RNN simultaneously. We utilize several tensor decompositions method including CANDECOMP/PARAFAC (CP), Tucker decomposition and Tensor Train (TT) to re-parameterize the Gated Recurrent Unit (GRU) RNN. We evaluate all tensor-based RNNs performance on sequence modeling tasks with a various number of parameters. Based on our experiment results, TT-GRU achieved the best results in a various number of parameters compared to other decomposition methods.
\end{abstract}

% no keywords

% For peer review papers, you can put extra information on the cover
% page as needed:
% \ifCLASSOPTIONpeerreview
% \begin{center} \bfseries EDICS Category: 3-BBND \end{center}
% \fi
%
% For peerreview papers, this IEEEtran command inserts a page break and
% creates the second title. It will be ignored for other modes.
\IEEEpeerreviewmaketitle

\section{Introduction}
% no \IEEEPARstart
In recent years, RNNs have achieved many state-of-the-arts on sequential data modeling task and significantly improved the performance on many tasks, such as speech recognition \cite{graves2013speech, graves2013hybrid} and machine translation \cite{bahdanau2014neural, sutskever2014sequence}. There are several reasons behind the RNNs impressive performance: the availability of data in large quantities and the advance of modern computer performances such as GPGPU. The recent hardware advance allows us to train and infer RNN models with million of parameters in a reasonable amount of time. 

Some devices such as mobile phones or embedded systems only have limited computing and memory resources. Therefore, deploying a model with a large number of parameters in those kind of devices is a challenging task. Therefore, we need to represent our model with more efficient methods and keep our model representational power at the same time. 

Some researchers have conducted important works to balance the trade-off between the model efficiency and their representational power. There are many different approaches to tackle this issue. From the low-level optimization perspective, Courbariaux et al. \cite{courbariaux2016binarized} replace neural network weight parameters with binary numbers. Hinton et al. \cite{hinton2015distilling} compress a larger model into a smaller model by training the latter on soft-target instead of hard-target. RNNs are composed by multiple linear transformations and followed by non-linear transformation. Most of RNN parameters are used to represent the weight matrix in those linear transformations and the total number of parameters depends on the input and hidden unit size. Therefore, some researchers also tried to represent the dense weight matrices with several alternative structures. Denil et al. \cite{denil2013predicting} employed low-rank matrix to replace the original weight matrix. 

Instead of using low-rank matrix decomposition, Novikov et al. \cite{novikov2015tensorizing} used TT format to represent the weight matrices in the fully connected layer inside a CNN model. Tjandra et al.\cite{tjandra2017compressing} applied TT-decomposition to compress the weight matrices inside RNN models. Besides TT-decomposition, there are several popular tensor decomposition methods such as CP decomposition and Tucker-decomposition.

However, those methods have not been explored for compressing RNN weight matrices, thus we are interested to see the extensive comparison between all tensor-decomposition method performances under the same number of parameters. In this paper, we utilized several tensor decomposition methods including CP-decomposition, Tucker decomposition and TT-decomposition for compressing RNN parameters. We represent GRU RNN weight matrices with these tensor decomposition methods. We conduct extensive experiments on sequence modeling with a polyphonic music dataset. We compare the performances of uncompressed GRU model and three different tensor-based compressed RNN models: CP-GRU, Tucker-GRU and TT-GRU \cite{tjandra2017compressing} on various number of parameters. From our experiment results, we conclude that TT-GRU achieved the best result in various number of parameters compared to other tensor-decomposition method.

\section{Recurrent Neural Network}
RNNs are a type of neural networks designed for modeling sequential and temporal data. For each timestep, the RNN calculate its hidden states by combining previous hidden states and a current input feature. Therefore, RNNs are able to capture all previous information from the beginning until current timestep. 

\subsection{Elman Recurrent Neural Network} \label{sec:elmanrnn}
\begin{figure}[h]
	\centering
	\includegraphics[width=6cm]{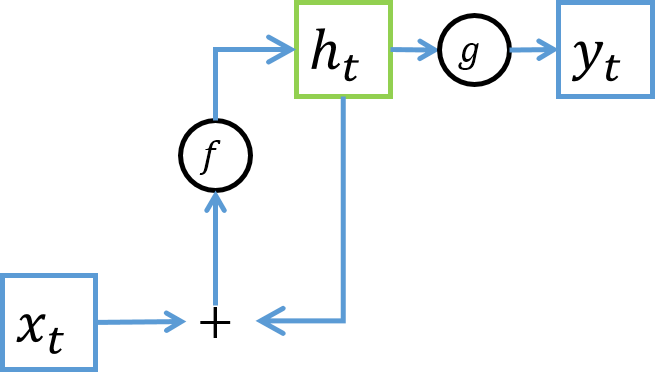}
	\caption{Recurrent Neural Network} 
	\label{fig:rnn}
\end{figure}

Elman RNNs are one of the earliest type of RNN models \cite{elman1990finding}. In some cases, Elman RNN also called as simple RNN.  Generally, we represent an input sequence as $\mathbf{x}=(x_1,...,x_{T})$, hidden vector sequence as $\mathbf{h}=(h_1,...,h_{T})$ and output vector sequence as $\mathbf{y}=(y_1,...,y_T)$. As illustrated in Fig.~\ref{fig:rnn}, a simple RNN at $t$-th time-step is can be formulated as:
\begin{eqnarray} \label{eq:simplernn}
h_t &=& f(W_{xh}x_t + W_{hh}h_{t-1}  + b_h) \\
y_t &=& g(W_{hy}h_t  + b_y). 
\end{eqnarray} where $W_{xh}$ represents the weight parameters between the input and hidden layer, $W_{hh}$ represents the weight parameters between the previous and current hidden layers, $W_{hy}$ represents the weight parameters between the hidden and output layer, and $b_h$ and $b_y$ represent bias vectors for the hidden and output layers. Functions $f(\cdot)$ and $g(\cdot)$ are nonlinear activation functions, such as sigmoid or tanh.
\subsection{Gated Recurrent Neural Network} \label{sec:gatedrnn}
Learning over long sequences is a hard problem for standard RNN because the gradient can easily vanish or explode \cite{bengio1994learning, hochreiter2001gradient}. One of the sources for that problem is because RNN equations are using bounded activation function such as tanh and sigmoid. Therefore, training a simple RNN is more complicated than training a feedforward neural network. Some researches addressed the difficulties of training simple RNNs. From the optimization perspective, Martens et al. \cite{martens2011learning} utilized a second-order Hessian-free (HF) optimization rather than the first-order method such as stochastic gradient descent. However, to calculate the second-order gradient or their approximation requires some extra computational steps. Le et al. \cite{le2015simple} changed the activation function that causes the vanishing gradient problem with a rectifier linear (ReLU) function. They are able to train a simple RNN for learning long-term dependency with an unbounded activation function and identity weight initialization. Modifying the internal structure from RNN by introducing gating mechanism also helps RNNs solve the vanishing gradient problems. The additional gating layers control the information flow from the previous states and the current input \cite{hochreiter1997long}. Several versions of gated RNNs have been designed to overcome the weakness of simple RNNs by introducing gating units, such as Long-Short Term Memory (LSTM) RNN and GRU RNN.

\subsubsection{Long-Short Term Memory RNN} \label{sec:lstm}
\begin{figure}[h]
	\centering
	\includegraphics[width=6.5cm]{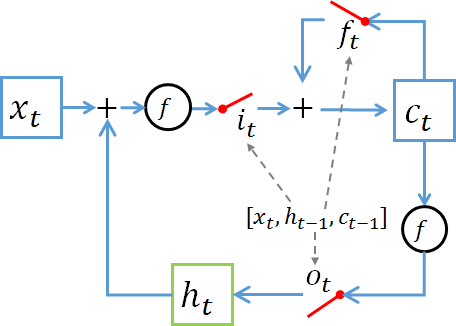}
	\caption{Long Short Term Memory Unit.}
	\label{fig:lstmrnn}
\end{figure}

An LSTM \cite{hochreiter1997long} is a gated RNN with memory cells and three gating layers. The gating layers purpose is to control the current memory states by retaining the important information and removing the unused information. The memory cells store the internal information across time steps. As illustrated in Fig.~\ref{fig:lstmrnn}, the LSTM hidden layer values at time $t$ are defined by the following equations \cite{graves2013hybrid}:
\begin{eqnarray}
i_t &=& \sigma(W_{xi} x_t + W_{hi} h_{t-1} + W_{ci} c_{t-1} + b_i) \\
f_t &=& \sigma(W_{xf} x_t + W_{hf} h_{t-1} + W_{cf} c_{t-1} + b_f) \\
c_t &=& f_t \odot c_{t-1} + i_t \odot \tanh(W_{xc} x_t + W_{hc} h_{t-1} + b_c) \\
o_t &=& \sigma(W_{xo} x_t + W_{ho} h_{t-1} + W_{co} c_t + b_o) \\
h_t &=& o_t \odot \tanh(c_t) 
\end{eqnarray}
where $\sigma(\cdot)$ is sigmoid activation function and $i_t, f_t, o_t$ and $c_t$ are respectively the input gates, the forget gates, the output gates and the memory cells. The input gates retain the candidate memory cell values that are useful for the current memory cell and the forget gates retain the previous memory cell values that are useful for the current memory cell. The output gates retain the memory cell values that are useful for the output and the next time-step hidden layer computation. 

\subsubsection{Gated Recurrent Unit RNN} \label{sec:gru}
\begin{figure}[h]
	\centering
	\includegraphics[width=6.5cm]{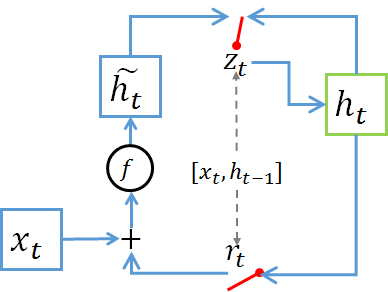}
	\caption{Gated Recurrent Unit}
	\label{fig:grurnn}
\end{figure}

A GRU \cite{cho2014learning} is one variant of gated RNN. It was proposed an alternative to LSTM. There are several key differences between GRU and LSTM. First, a GRU does not seperate the hidden states with the memory cells \cite{chung2014empirical}. Second, instead of three gating layers, it only has two: reset gates and update gates.
As illustrated in Fig.~\ref{fig:grurnn}, the GRU hidden layer at time $t$ is defined by the following equations \cite{cho2014learning}:
\begin{eqnarray}
r_t &=& \sigma(W_{xr} x_t + W_{hr} h_{t-1} + b_r) \label{eq:grureset} \\
z_t &=& \sigma(W_{xz} x_t + W_{hz} h_{t-1} + b_z) \label{eq:gruupdate} \\
\tilde{h_t} &=& f(W_{xh} x_t + W_{hh} (r_t \odot h_{t-1}) + b_h) \label{eq:grucandidate}\\
h_t &=& (1 - z_t) \odot h_{t-1} + z_t \odot \tilde{h_t} \label{eq:gruhidden}
\end{eqnarray}
where $\sigma(\cdot)$ is a sigmoid activation function, $f(\cdot)$ is a tanh activation function, $r_t, z_t$ are the reset and update gates, $\tilde{h_t}$ is the candidate hidden layer values, and $h_t$ is the hidden layer values at time $t$-th. The reset gates control the previous hidden layer values that are useful for the current candidate hidden layer. The update gates decide whether to keep the previous hidden layer values or replace the current hidden layer values with the candidate hidden layer values. 
GRU can match LSTM's performance and its convergence speed sometimes surpasses LSTM, despite having one fewer gating layer \cite{chung2014empirical}.

\section{Tensor RNN}
In this section, we explain our approaches to compress the parameters in the RNN. First, we define the tensorization process to transform the weight matrices inside the RNN model into higher order tensors. Then, we describe two tensor decompositions method called as CANDECOMP/PARAFAC (CP) decomposition and Tucker decomposition. Last, we explain about tensorization and RNN parameters compression with the tensor decomposition methods.

\subsection{Vector, Matrix and Tensor}
Before we start to explain any further, we will define different notations for vectors, matrices and tensors. Vector is an one-dimensional array, matrix is a two-dimensional array and tensor is a higher-order multidimensional array. 
In this paper, bold lower case letters (e.g., $\mathbf{b}$) represent vectors, bold upper case letters (e.g., $\mathbf{W}$) represent matrices and bold calligraphic upper case letters (e.g., $\mathbfcal{W}$) represent tensors. For representing the element inside vectors, matrices and tensors, we explicitly write the index in every dimension without bold font. For example, $b(i)$ is the $i$-th element in vector $\mathbf{b}$, $W(p, q)$ is the element on $p$-th row and $q$-th column from matrix $\mathbf{W}$ and $\mathcal{W}(i_1,..,i_d)$ is the $i_1,..,i_d$-th index from tensor $\mathbfcal{W}$.

\subsection{Tensor decomposition method}
Tensor decomposition is a method for generalizing low-rank approximation from a multi-dimensional array. There are several popular tensor decomposition methods, such as Canonical polyadic (CP) decomposition, Tucker decomposition and Tensor Train decomposition. The factorization format differs across different decomposition methods. In this section, we explain briefly about CP-decomposition and Tucker decomposition.
\subsubsection{CP-decomposition} \label{sec:cpdecom}
\begin{figure}[]
	\centering
	\includegraphics[width=7.5cm]{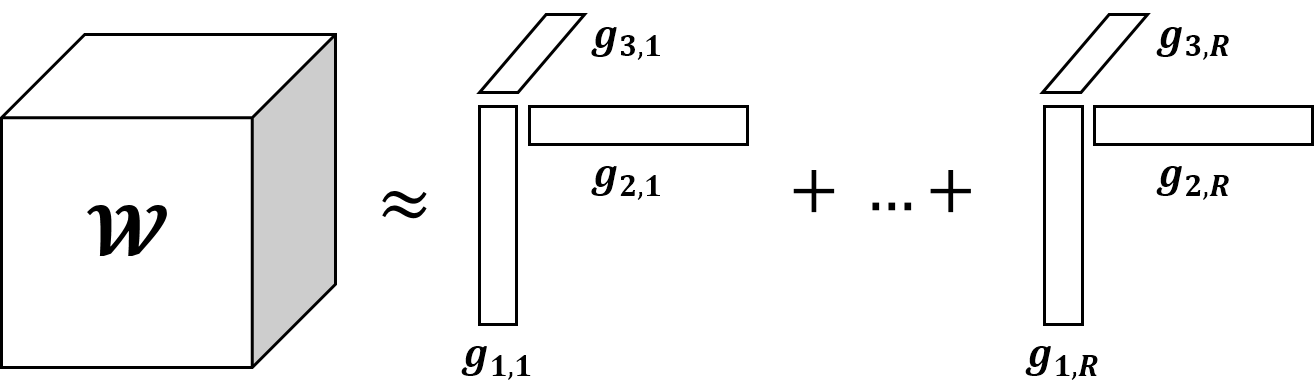}
	\caption{CP-decomposition for 3rd-order tensor $\mathbfcal{W}$}
	\label{fig:cpdecom}
\end{figure}

Canonical polyadic decomposition (CANDECOMP/PARAFAC) \cite{harshman1970foundations, kiers2000towards, kolda2009tensor} or usually referred to CP-decomposition factorizes a tensor into the sum of outer products of vectors. Assume we have a 3rd-order tensor $\mathbfcal{W} \in \mathbb{R}^{m_1 \times m_2 \times m_3}$, we can approximate it with CP-decomposition:
\begin{eqnarray}
	\mathbfcal{W} \approx \sum_{r=1}^{R} \mathbf{g}_{1,r} \otimes \mathbf{g}_{2,r} \otimes \mathbf{g}_{3,r} \label{eq:cpdecom}
\end{eqnarray} 
where $\forall r\in[1..R],  \mathbf{g}_{1,r} \in \mathbb{R}^{m_1}$, $\mathbf{g}_{2,r} \in \mathbb{R}^{m_2}$, $\mathbf{g}_{3,r} \in \mathbb{R}^{m_3}$, $R \in \mathbb{Z}^{+}$ is the number of factors combinations (CP-rank) and $\otimes$ denotes Kronecker product operation. Elementwise, we can calculate the result by:
\begin{eqnarray}
	\mathcal{W}(x,y,z) \approx \sum_{r=1}^{R} g_{1,r}(x) 
	\, g_{2,r}(y) \, g_{3,r}(z) 
\end{eqnarray}
In Figure~\ref{fig:cpdecom}, we provide an illustration for Eq.~\ref{eq:cpdecom} in more details.

\subsubsection{Tucker decomposition}\label{sec:tuckerdecom}
\begin{figure}[h]
	\centering
	\includegraphics[width=7.5cm]{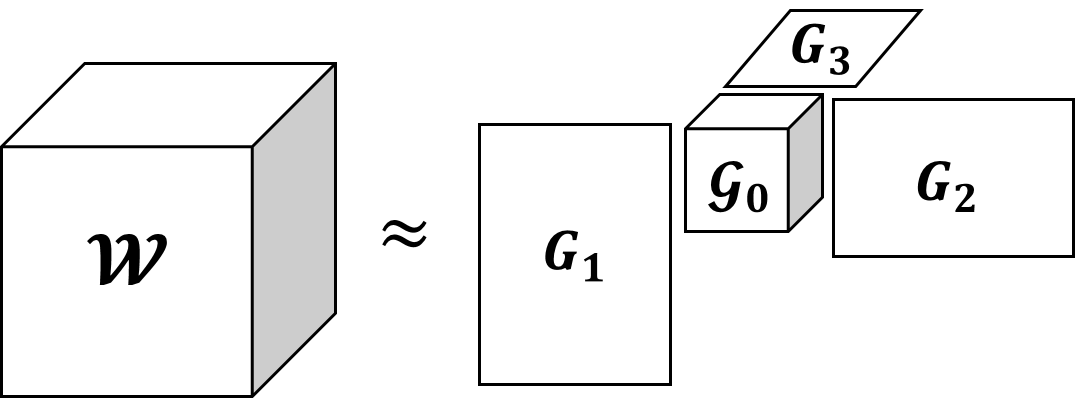}
	\caption{Tucker decomposition for 3rd-order tensor $\mathbfcal{W}$}
	\label{fig:tuckerdecom}
\end{figure}

Tucker decomposition \cite{tucker1966some, kolda2009tensor} factorizes a tensor into a core tensor multiplied by a matrix along each mode. Assume we have a 3rd-order tensor $\mathbfcal{W} \in \mathbb{R}^{m_1 \times m_2 \times m_3}$, we can approximate it with Tucker decomposition:
\begin{eqnarray}
	\mathbfcal{W} \approx \mathbfcal{G}_0 \times_1 \mathbf{G_{1}} \times_2 \mathbf{G}_2 \times_2 \mathbf{G}_3 \label{eq:tuckerdecom}
\end{eqnarray} 
where $\mathbfcal{G}_0 \in \mathbb{R}^{r_1 \times r_2 \times r_3}$ is the core tensor, $\mathbf{G}_1 \in \mathbb{R}^{m_1 \times r_1}$, $\mathbf{G}_2 \in \mathbb{R}^{m_2 \times r_2}$, $\mathbf{G}_3 \in \mathbb{R}^{m_3 \times r_3}$ are the factor matrices and $\times_n$ is the n-th mode product operator. The mode product between a tensor $\mathbfcal{G}_0 \in \mathbb{R}^{n_1 \times n_2 \times n_3}$ and a matrix $\mathbf{G}_1 \in \mathbb{R}^{m_1 \times n_1}$ is a tensor $\mathbb{R}^{m_1 \times n_2 \times n_3}$. By applying the mode products across all modes, we can recover the original $\mathbfcal{W}$ tensor. Elementwise, we can calculate the element from tensor $\mathbfcal{W}$ by:
\begin{align}
\mathcal{W}(x,y,z) \approx \nonumber & \sum_{s_1=1}^{r_1}\sum_{s_2=1}^{r_2}\sum_{s_3=1}^{r_3} \mathcal{G}_0(s_1, s_2, s_3) \\
& \, G_1(x, s_1) \, G_2(y, s_2) \, G_3(z, s_3) 
\end{align} where $x \in [1,..,m_1], y \in [1,..,m_2], z \in [1,.., m_3]$. Figure~\ref{fig:tuckerdecom} gives an illustration for Eq.~\ref{eq:tuckerdecom}

\subsection{Tensor Train decomposition}
\begin{figure}[h]
	\centering
	\includegraphics[width=7.5cm]{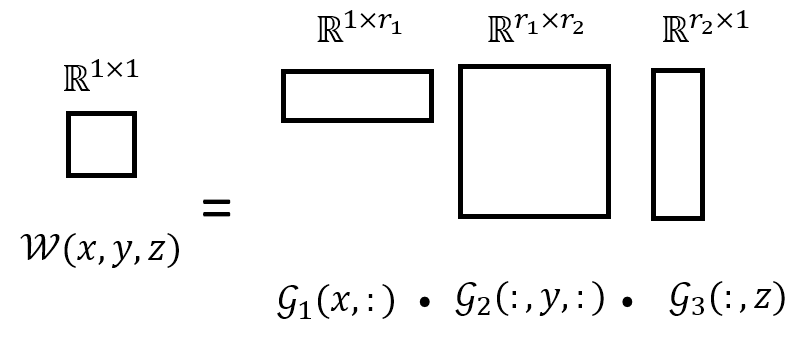}
	\caption{Tensor Train decomposition for 3rd-order tensor $\mathbfcal{W}$}
	\label{fig:tensortrain}
\end{figure}

Tensor Train decomposition \cite{oseledets2011tt} factorizes a tensor into a collection of lower order tensors called as TT-cores. All TT-cores are connected through matrix multiplications across all tensor order to calculate the element from original tensor. Assume we have a 3rd-order tensor $\mathbfcal{W} \in \mathbb{R}^{m_1 \times m_2 \times m_3}$, we can approximate the element at index $x, y, z$ by:
\begin{align}
\mathbfcal{W}(x, y, z) \approx \sum_{s_1=1}^{r_1} \sum_{s_2=1}^{r_2} \mathbfcal{G}_{1}(x, s_1) \mathbfcal{G}_{2}(s_1, y, s_2) \mathbfcal{G}_{3}(s_2, z) \label{eq:tensortrain}
\end{align} where $x \in [1,..,m_1], y \in [1,..,m_2], z \in [1,.., m_3]$ and $\mathbfcal{G}_1 \in \mathbb{R}^{m_1 \times r_1}, \mathbfcal{G}_2 \in \mathbb{R}^{r_1 \times m_2 \times r_2}, \mathbfcal{G}_3 \in \mathbb{R}^{r_2 \times m_3}$ as the TT-cores. Figure \ref{fig:tensortrain} gives an illustration for Eq.~\ref{eq:tensortrain}.

\subsection{RNN parameters tensorization}
Most of RNN equations are composed by multiplication between the input vector and their corresponding weight matrix: 
\begin{eqnarray}
	\mathbf{y} = \mathbf{W} \mathbf{x} + \mathbf{b} \label{eq:linproj}
\end{eqnarray} where $\mathbf{W} \in \mathbb{R}^{M \times N}$ is the weight matrix, $\mathbf{b} \in \mathbb{R}^{M}$ is the bias vector and $\mathbf{x} \in \mathbb{R}^{N}$ is the input vector. Thus, most of RNN parameters are used to represent the weight matrices. To reduce the number of parameters significantly, we need to represent the weight matrices with the factorization of higher-order tensor. 
First, we apply tensorization on the weight matrices. Tensorization is the process to transform a lower-order dimensional array into a higher-order dimensional array. In our case, we tensorize RNN weight matrices into tensors. Given a weight matrix $\mathbf{W} \in \mathbb{R}^{M \times N}$, we can represent them as a tensor $\mathbfcal{W} \in \mathbb{R}^{m_1 \times m_2 \times .. \times m_d \, \times n_1 \times n_2 \times .. \times n_d}$ where $M=\prod_{k=1}^{d}{m_k}$ and $N=\prod_{k=1}^{d}{n_k}$. For mapping each element in matrix $\mathbf{W}$ to tensor $\mathbfcal{W}$, we define one-to-one mapping between row-column and tensor index with bijective functions $\mathbf{f}_i: \mathbb{Z}_{+} \rightarrow \mathbb{Z}_{+}^{d}$ and $\mathbf{f}_j : \mathbb{Z}_{+} \rightarrow \mathbb{Z}_{+}^{d}$. Function $\mathbf{f}_i$ transforms each row $p \in \{1,..,M\}$ into $\mathbf{f}_i(p) = [i_1(p),..,i_d(p)]$ and $\mathbf{f}_j$ transforms each column $q \in \{1,..,N\}$ into $\mathbf{f}_j(q)=[j_1(q),..,j_d(q)]$. Following this, we can access the value from matrix $W(p, q)$ in the tensor $\mathbfcal{W}$ with the index vectors generated by $\mathbf{f}_i(p)$ and $\mathbf{f}_j(q)$ with these bijective functions.

After we determine the shape of the weight tensor, we choose one of the tensor decomposition methods (e.g., CP-decomposition (Sec.\ref{sec:cpdecom}), Tucker decomposition (Sec.\ref{sec:tuckerdecom}) or Tensor Train\cite{oseledets2011tt}) to represent and reduce the number of parameters from the tensor $\mathbfcal{W}$. In order to represent matrix-vector products inside RNN equations, we need to reshape the input vector $\mathbf{x} \in \mathbb{R}^{N}$ into a tensor $\mathbfcal{X} \in \mathbb{R}^{n_1 \times .. \times n_d}$ and the bias vector $\mathbf{b} \in \mathbb{R}^{M}$ into a tensor $\mathbfcal{B} \in \mathbb{R}^{m_1 \times .. \times m_d}$. Therefore, we can reformulate the Eq.~\ref{eq:linproj} to calculate $y(p)$  elementwise with:
\begin{align}
\mathcal{Y}(\mathbf{f}_i(p)) = \sum_{j_1,..,j_d}^{} & \mathcal{W}\left(\mathbf{f}_i(p), j_1,..,j_d\right) \, \mathcal{X}(j_1,..,j_d) \nonumber \\ 
& + \mathcal{B}(\mathbf{f}_i(p)) \label{eq:tensorproj}
\end{align} by enumerating all columns $q$ position with $j_1,..,j_d$ and $\mathbf{f}_i(p) = [i_1(p),..,i_d(p)]$. 

For CP-decomposition, we represent our tensor $\mathbfcal{W}$ with multiple factors $\mathbf{gm}_{k,r}, \mathbf{gn}_{k, r}$ where $\forall k\in[1..d]\,\forall r\in[1..R], \, (\mathbf{gm}_{k,r} \in \mathbb{R}^{m_k}, \mathbf{gn}_{k,r} \in \mathbb{R}^{n_k})$. From here, we replace Eq.~\ref{eq:tensorproj} with:
\begin{align}
\mathcal{Y}(\mathbf{f}_i(p)) = \sum_{j_1,..,j_d}^{} & 
\left(\sum_{r=1}^{R}\prod_{k=1}^{d} {gm}_{k, r}(i_k(p)) gn_{k,r}(j_k) \right) \, \mathcal{X}(j_1,..,j_d) \nonumber \\ 
& + \mathcal{B}(\mathbf{f}_i(p)) \label{eq:cptensorproj} .
\end{align} 
By using CP-decomposition for representing the weight matrix $\mathbf{W}$, we reduce the number of parameters from $M \times N$ into $R * (\sum_{k=1}^{d} m_k+n_k)$.

For Tucker decomposition, we represent out tensor $\mathbfcal{W}$ with a tensor core $\mathbfcal{G}_0 \in \mathbb{R}^{r_1 \times ... \times r_d \times r_{d+1} \times ... \times r_{2d}}$ where $\forall k\in [1..d], \, r_k < m_k$ and $\forall k\in [1..d], \, r_{d+k} < n_k $ and multiple factor matrices $\mathbf{GM}_k, \mathbf{GN}_k$, where $\forall k \in [1..d],\, (\mathbf{GM}_k \in \mathbb{R}^{m_k \times r_{k}}, \mathbf{GN}_k \in \mathbb{R}^{n_k \times r_{d+k}})$. Generally, the tensor core ranks $r_1, r_2,..,r_d$ are corresponding to the row in tensor index and $r_{d+1}, r_{d+2},..,r_{2d}$ are corresponding to the column in tensor index. From here, we replace Eq.~\ref{eq:tensorproj} with:
\begin{equation}
\begin{split}
\mathcal{Y}(\mathbf{f}_i(p)) = \sum_{j_1,..,j_d}^{}  
&\left( \sum_{s_1,..s_d, s_{d+1},..,s_{2d}}^{r_1,..,r_d,r_{d+1},..,r_{2d}} \mathcal{G}_0(s_1,..s_d, s_{d+1},..,s_{2d}) \right. \\
&\left. \prod_{k=1}^{d}  GM_k(i_k(p), s_k) GN_k(j_k, s_{d+k}) \right) \, \mathcal{X}(j_1,..,j_d) \\ 
&+ \mathcal{B}(\mathbf{f}_i(p)) \label{eq:tuckertensorproj} .
\end{split}
\end{equation}
By using Tucker decomposition for representing the weight matrix $\mathbf{W}$, we reduce the number of parameters from $M \times N$ into $\sum_{k=1}^{d} (m_k * r_k +  n_k * r_{d+k})+ (\prod_{k=1}^{2d}r_k) $.

For the TT-decomposition, we refer to \cite{tjandra2017compressing} on how to represent the tensor $\mathbfcal{W}$ and how to calculate the linear projection to replace Eq.~\ref{eq:tensorproj}.

In this work, we focus on compressing GRU-RNN by representing all weight matrices (input-to-hidden and hidden-to-hidden) with tensors and factorize the tensors with low-rank tensor decomposition methods. For compressing other RNN architectures such as Elman RNN (Sec.~\ref{sec:elmanrnn}) or LSTM-RNN (Sec.~\ref{sec:lstm}), we can follow the same steps by replacing all the weight matrices with factorized tensors representation.

\subsection{Tensor Core and Factors Initialization Trick}
Because of the large number of recursive matrix multiplications, followed by some nonlinearity (e.g, sigmoid, tanh), the gradient from the hidden layer will diminish after several time-step \cite{pascanu2013difficulty}. Consequently, training recurrent neural networks is much harder compared to standard feedforward neural networks. 

Even worse, we decompose the weight matrix into multiple smaller tensors or matrices, thus the number of multiplications needed for each calculation increases multiple times. Therefore, we need a better initialization trick on the tensor cores and factors to help our model convergences in the early training stage.

In this work, we follow Glorot et al.\cite{glorot2010understanding} by initializing the weight matrix with a certain variance. We assume that our original weight matrix $\mathbf{W}$ has a mean $0$ and the variance $\sigma_W^2$. We utilize the basic properties from a sum and a product variance between two independent random variables.
\begin{definition} \label{def:sumvar}
	Let $X$ and $Y$ be independent random variables with the mean $0$, then the variance from the sum of $X$ and $Y$ is $Var(X + Y) = Var(X) + Var(Y)$ 
\end{definition}

\begin{definition} \label{def:prodvar}
	Let $X$ and $Y$ be independent random variables with the mean $0$, then the variance from the product of $X$ and $Y$ is $Var(X * Y) = Var(X) * Var(Y)$
\end{definition}
After we decided the target variance $\sigma_w^2$ for our original weight matrix, now we need to derive the proper initialization rules for the tensor core and factors. We calculate the variance for tensor core and factors by observing the number of sum and product operations and utilize the variance properties from Def.~\ref{def:sumvar} and \ref{def:prodvar}.
For weight tensor $\mathbfcal{W}$ based on the CP-decomposition, we can calculate $\sigma_{g}$ as the standard deviation for all factors $\mathbf{gm}_{k,r}, \mathbf{gn}_{k,r}$ with:
\begin{eqnarray}
\sigma_{g}=\sqrt[\leftroot{-2}\uproot{2} 4d]{\frac{\sigma_{w}^2}{R}}
\end{eqnarray} and initialize $\mathbf{gm}_{k,r}, \mathbf{gn}_{k,r} \sim \mathcal{N}(0, \sigma_{g}^2)$.

For weight tensor $\mathbfcal{W}$ based on the Tucker decomposition, we can calculate $\sigma_{g}$ as the standard deviation for the core tensor $\mathbfcal{G}_0$ and the factor matrices $\mathbf{GM}_k, \mathbf{GN}_k$ with:
\begin{eqnarray}
\sigma_{g} = \sqrt[\leftroot{-2}\uproot{4} (4d+2)]{ \frac{\sigma_w^2}{\prod_{k=1}^{2d}r_k}}
\end{eqnarray} and initialize $\mathbfcal{G}_0, \mathbf{GM}_{k}, \mathbf{GN}_{k} \sim \mathcal{N}(0, \sigma_{g}^2)$.

For weight tensor $\mathbfcal{W}$ based on the Tensor Train decomposition, we refer to \cite{tjandra2017compressing} for initializing the TT-cores $\mathbfcal{G}_i$.

\section{Experiments}
In this section, we describe our dataset and all model configurations. We performed experiments with three different tensor-decompositions (CP decomposition, Tucker decomposition and TT decomposition) to compress our GRU and also the baseline GRU. In the end, we report our experiment results and finish this section with some discussions and conclusions. Our codes are available at \texttt{\color{blue} \url{https://github.com/androstj/tensor_rnn}}.

\subsection{Dataset}
We evaluated our models with sequential modeling tasks. We used a polyphonic music dataset \cite{boulanger2012} which contains 4 different datasets\footnote{Dataset are downloaded from: \url{http://www-etud.iro.umontreal.ca/~boulanni/icml2012}}: Nottingham, MuseData, PianoMidi and JSB Chorales. For each active note in all time-step, we set the value as 1, otherwise 0. Each dataset consists of at least 7 hours of polyphonic music and the total is $\pm$ 67 hours. 
\subsection{Models}
We evaluate several models in this paper: GRU-RNN (no compression), CP-GRU (weight compression via CP decomposition), Tucker-GRU (weight compression via Tucker decomposition), TT-GRU \cite{tjandra2017compressing} (compressed weight with TT-decomposition).
For each timestep, the input and output targets are vectors of 88 binary value. The input vector is projected by a linear layer with 256 hidden units, followed by LeakyReLU\cite{maas2013rectifier} activation function. For the RNN model configurations, we enumerate all the details in the following list:
\begin{enumerate}
	\item GRU
		\begin{itemize}
			\item Input size ($N$): 256
			\item Hidden size ($M$): 512
		\end{itemize}
	\item Tensor-based GRU
		\begin{itemize}
			\item Input size ($N$): 256
			\item Tensor input shape ($n_{1..4}$): $4 \times 4 \times 4 \times 4$
			\item Hidden size ($M$): 512
			\item Tensor hidden shape ($m_{1..4}$): $8 \times 4 \times 4 \times 4$
		\end{itemize}
	\begin{enumerate}
		\item CP-GRU
		\begin{itemize}
			\item CP-Rank ($R$): $[10, 30, 50, 80, 110]$
		\end{itemize}
		\item Tucker-GRU
		\begin{itemize}
			\item Core ($\mathbfcal{G}_0$) shape: 
			\begin{itemize}
				\item $(2 \times 2 \times 2 \times 2) \times (2 \times 2 \times 2 \times 2)$
				\item $(2 \times 3 \times 2 \times 3) \times (2 \times 3 \times 2 \times 3)$
				\item $(2 \times 3 \times 2 \times 4) \times (2 \times 3 \times 2 \times 4)$
				\item $(2 \times 4 \times 2 \times 4) \times (2 \times 4 \times 2 \times 4)$
				\item $(2 \times 3 \times 3 \times 4) \times (2 \times 3 \times 3 \times 4)$
			\end{itemize}
		\end{itemize}
		
		\item TT-GRU
		\begin{itemize}
			\item TT-ranks:
			\begin{itemize}
				\item $(1 \times 3 \times 3 \times 3 \times 1)$
				\item $(1 \times 5 \times 5 \times 5 \times 1)$
				\item $(1 \times 7 \times 7 \times 7 \times 1)$
				\item $(1 \times 9 \times 9 \times 9 \times 1)$
				\item $(1 \times 9 \times 9 \times 9 \times 1)$
			\end{itemize}
		\end{itemize}
	\end{enumerate}	
\end{enumerate}

In this task, the training criterion is to minimize the negative log-likelihood (NLL). In evaluation, we measured two different scores: NLL and accuracy (ACC). For calculating the accuracy, we follow Bay et al. \cite{bay2009evaluation} formulation: 
\begin{equation} 
ACC = \frac{\sum_{t=1}^T TP(t)}{\sum_{t=1}^T \left(TP(t)+FP(t)+FN(t)\right)}
\end{equation} where $TP(t), FP(t), FN(t)$ is the true positive, false positive and false negative at time-$t$. 

For training models, we use Adam \cite{kingma2014adam} algorithm for our optimizer. To stabilize our training process, we clip our gradient when the norm $\left|\left| \nabla w \right|\right| \, > 5$. For fair comparisons, we performed a grid search over learning rates ($1e-2, 5e-3, 1e-3$) and dropout probabilities ($0.2, 0.3, 0.4, 0.5$). The best model based on loss in validation set will be used for the test set evaluation.

\subsection{Result and Discussion}
	
% Please add the following required packages to your document preamble:
% \usepackage{multirow}
% Please add the following required packages to your document preamble:
% \usepackage{multirow}
% \usepackage[table,xcdraw]{xcolor}
% If you use beamer only pass "xcolor=table" option, i.e. \documentclass[xcolor=table]{beamer}
\begin{table*}[t]
	\centering
	\caption{Comparison between all models and their configurations based on the number of parameters, negative log-likelihood and accuracy of polyphonic test set}
	\label{tbl:allexp}
	\begin{tabular}{|c|c|l|l|l|l|l|l|l|l|l|}
		\hline
		&                                   & \multicolumn{1}{c|}{}                                 & \multicolumn{8}{c|}{\textbf{Dataset}}                                                                                                                                                                                                                                                                                                                                                                         \\ \cline{4-11} 
		&                                   & \multicolumn{1}{c|}{}                                 & \multicolumn{2}{c|}{\textbf{Nottingham}}                                                          & \multicolumn{2}{c|}{\textbf{JSB}}                                                                 & \multicolumn{2}{c|}{\textbf{PianoMidi}}                                                           & \multicolumn{2}{c|}{\textbf{MuseData}}                                                            \\ \cline{4-11} 
		\multirow{-3}{*}{\textbf{Model}}                                                                  & \multirow{-3}{*}{\textbf{Config}} & \multicolumn{1}{c|}{\multirow{-3}{*}{\textbf{Param}}} & \textbf{NLL}                                    & \textbf{ACC}                                    & \textbf{NLL}                                    & \textbf{ACC}                                    & \textbf{NLL}                                    & \textbf{ACC}                                    & \textbf{NLL}                                    & \textbf{ACC}                                    \\ \hline
		\begin{tabular}[c]{@{}c@{}}GRU\\ IN:256\\ OUT:512\end{tabular}                                    &                                   & 1181184                                               & 3.369 & 71.1 & 8.32 & 30.24 & 7.53 & 27.19 & 7.12 & 36.30 \\ \hline
		& \textbf{Rank}                     & \cellcolor[HTML]{9B9B9B}{\color[HTML]{9B9B9B} }       & \cellcolor[HTML]{9B9B9B}{\color[HTML]{9B9B9B} } & \cellcolor[HTML]{9B9B9B}{\color[HTML]{9B9B9B} } & \cellcolor[HTML]{9B9B9B}{\color[HTML]{9B9B9B} } & \cellcolor[HTML]{9B9B9B}{\color[HTML]{9B9B9B} } & \cellcolor[HTML]{9B9B9B}{\color[HTML]{9B9B9B} } & \cellcolor[HTML]{9B9B9B}{\color[HTML]{9B9B9B} } & \cellcolor[HTML]{9B9B9B}{\color[HTML]{9B9B9B} } & \cellcolor[HTML]{9B9B9B}{\color[HTML]{9B9B9B} } \\ \cline{2-11} 
		& 10                                & 2456                                                  & 3.79                                            & 67.51                                           & 8.60                                            & 27.29                                           & 8.15                                            & 19.03                                           & 7.87                                            & 27.32                                           \\ \cline{2-11} 
		& 30                                & 4296                                                  & 3.48                                            & 69.85                                           & 8.49                                            & 28.33                                           & 7.68                                            & 25.03                                           & 7.27                                            & 36.19                                           \\ \cline{2-11} 
		& 50                                & 6136                                                  & 3.46                                            & 69.56                                           & 8.40                                            & 28.47                                           & 7.66                                            & 26.18                                           & 7.23                                            & 36.34                                           \\ \cline{2-11} 
		& 80                                & 8896                                                  & 3.43                                            & 69.73                                           & 8.41                                            & 27.88                                           & 7.61                                            & 28.28                                           & 7.19                                            & 36.57                                           \\ \cline{2-11} 
		\multirow{-6}{*}{\begin{tabular}[c]{@{}c@{}}CP-GRU\\ IN: 4,4,4,4\\ OUT: 8,4,4,4\end{tabular}}     & 110                               & 11656                                                 & 3.34                                            & 70.42                                           & 8.41                                            & 29.45                                           & 7.60                                            & 27.36                                           & 7.18                                            & 36.89                                           \\ \hline
		& \textbf{Cores}                    & \cellcolor[HTML]{9B9B9B}{\color[HTML]{9B9B9B} }       & \cellcolor[HTML]{9B9B9B}{\color[HTML]{9B9B9B} } & \cellcolor[HTML]{9B9B9B}{\color[HTML]{9B9B9B} } & \cellcolor[HTML]{9B9B9B}{\color[HTML]{9B9B9B} } & \cellcolor[HTML]{9B9B9B}{\color[HTML]{9B9B9B} } & \cellcolor[HTML]{9B9B9B}{\color[HTML]{9B9B9B} } & \cellcolor[HTML]{9B9B9B}{\color[HTML]{9B9B9B} } & \cellcolor[HTML]{9B9B9B}{\color[HTML]{9B9B9B} } & \cellcolor[HTML]{9B9B9B}{\color[HTML]{9B9B9B} } \\ \cline{2-11} 
		& 2,2,2,2                           & 2232                                                  & 3.71                                            & 68.30                                           & 8.57                                            & 27.28                                           & 7.98                                            & 20.79                                           & 7.81                                            & 29.94                                           \\ \cline{2-11} 
		& 2,3,2,3                           & 4360                                                  & 3.64                                            & 68.63                                           & 8.48                                            & 28.10                                           & 7.75                                            & 24.92                                           & 7.38                                            & 34.20                                           \\ \cline{2-11} 
		& 2,3,2,4                           & 6408                                                  & 3.55                                            & 69.10                                           & 8.44                                            & 28.06                                           & 7.73                                            & 25.66                                           & 7.69                                            & 32.50                                           \\ \cline{2-11} 
		& 2,4,2,4                           & 10008                                                 & 3.52                                            & 69.18                                           & 8.41                                            & 27.70                                           & 7.75                                            & 24.46                                           & 7.38                                            & 35.58                                           \\ \cline{2-11} 
		\multirow{-6}{*}{\begin{tabular}[c]{@{}c@{}}TUCKER-GRU\\ IN: 4,4,4,4\\ OUT: 8,4,4,4\end{tabular}} & 2,3,3,4                           & 12184                                                 & 3.41                                            & 70.23                                           & 8.43                                            & 29.03                                           & 7.69                                            & 25.26                                           & 7.43                                            & 33.63                                           \\ \hline
		& \textbf{TT-rank}                  & \cellcolor[HTML]{9B9B9B}{\color[HTML]{9B9B9B} }       & \cellcolor[HTML]{9B9B9B}{\color[HTML]{9B9B9B} } & \cellcolor[HTML]{9B9B9B}{\color[HTML]{9B9B9B} } & \cellcolor[HTML]{9B9B9B}{\color[HTML]{9B9B9B} } & \cellcolor[HTML]{9B9B9B}{\color[HTML]{9B9B9B} } & \cellcolor[HTML]{9B9B9B}{\color[HTML]{9B9B9B} } & \cellcolor[HTML]{9B9B9B}{\color[HTML]{9B9B9B} } & \cellcolor[HTML]{9B9B9B}{\color[HTML]{9B9B9B} } & \cellcolor[HTML]{9B9B9B}{\color[HTML]{9B9B9B} } \\ \cline{2-11} 
		& 1,3,3,3,1                         & 2688                                                  & 3.49                                            & 69.49                                           & 8.37                                            & 28.41                                           & 7.60                                            & 26.95                                           & 7.49                                            & 34.99                                           \\ \cline{2-11} 
		& 1,5,5,5,1                         & 4096                                                  & 3.45                                            & 69.81                                           & 8.38                                            & 28.86                                           & 7.58                                            & 27.46                                           & 7.50                                            & 33.37                                           \\ \cline{2-11} 
		& 1,7,7,7,1                         & 6016                                                  & 3.40                                            & 70.72                                           & 8.37                                            & 28.83                                           & 7.57                                            & 27.58                                           & 7.23                                            & 36.53                                           \\ \cline{2-11} 
		& 1,9,9,9,1                         & 8448                                                  & 3.35                                            & 70.82                                           & 8.36                                            & 29.32                                           & 7.58                                            & 27.62                                           & 7.20                                            & 37.81                                           \\ \cline{2-11} 
		\multirow{-6}{*}{\begin{tabular}[c]{@{}c@{}}TT-GRU\\ IN: 4,4,4,4\\ OUT: 8,4,4,4\end{tabular}}     & 1,11,11,11,1                      & 11392                                                 & 3.38                                            & 70.51                                           & 8.37                                            & 29.55                                           & 7.58                                            & 28.07                                           & 7.16                                            & 36.54                                           \\ \hline
	\end{tabular}
\end{table*}

%\begin{figure}[h]
%	\centering
%	\includegraphics[width=0.45\textwidth]{img/plot_alldata_portrait.png}
%	\caption{Negative log-likelihood (NLL) comparison between GRU, TT-GRU, Tucker-GRU, and CP-GRU on polyphonic test set}
%	\label{fig:plot_alldata}
%\end{figure}

%
\begin{figure}[h!]
	\centering
	\includegraphics[width=0.45\textwidth]{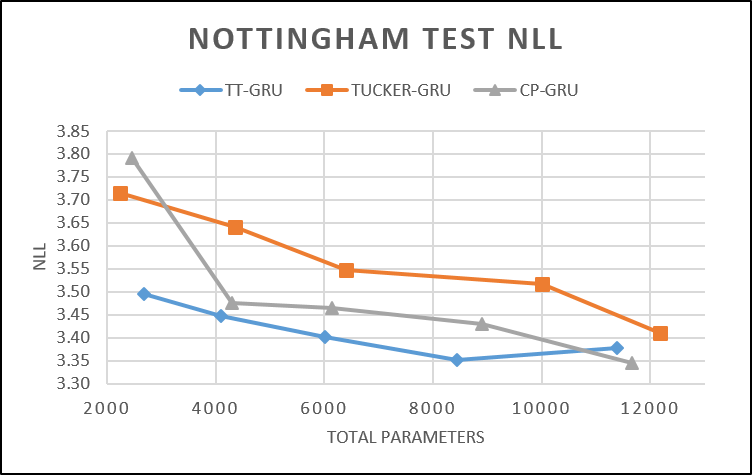}
	\caption{NLL comparison between TT-GRU, Tucker-GRU, and CP-GRU on Nottingham test set} 
	\label{fig:plot_nottingham}
\end{figure}

\begin{figure}[h!]
	\centering
	\includegraphics[width=0.45\textwidth]{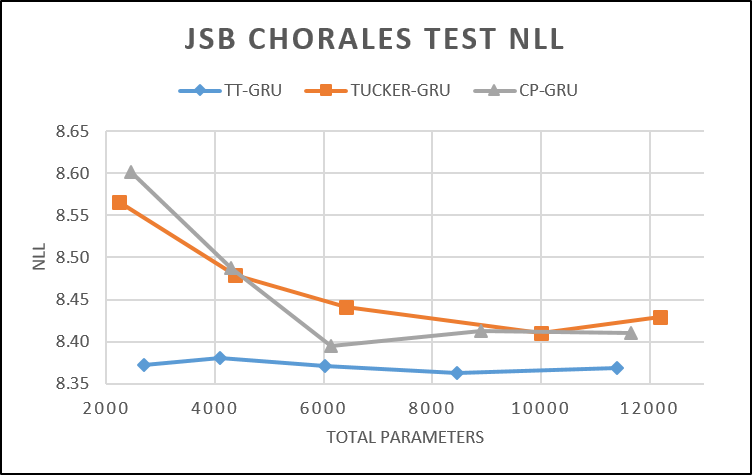}
	\caption{NLL comparison between TT-GRU, Tucker-GRU, and CP-GRU on JSB Chorales test set} 
	\label{fig:plot_jsb}
\end{figure}

\begin{figure}[h!]
	\centering
	\includegraphics[width=0.45\textwidth]{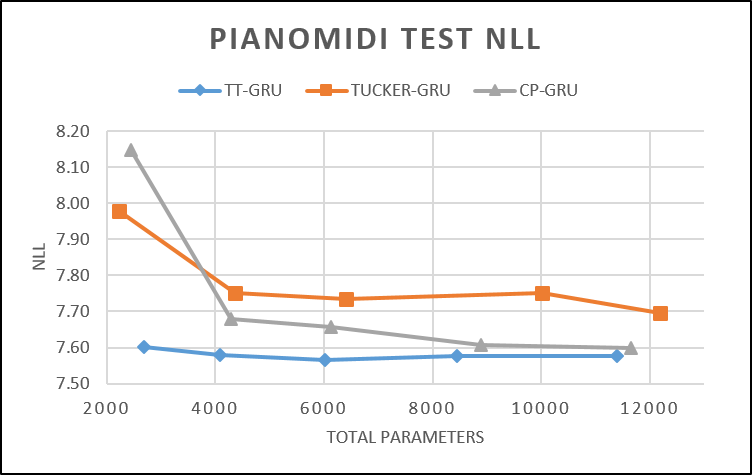}
	\caption{NLL comparison between TT-GRU, Tucker-GRU, and CP-GRU on PianoMidi test set} 
	\label{fig:plot_pianomidi}
\end{figure}

\begin{figure}[h!]
	\centering
	\includegraphics[width=0.45\textwidth]{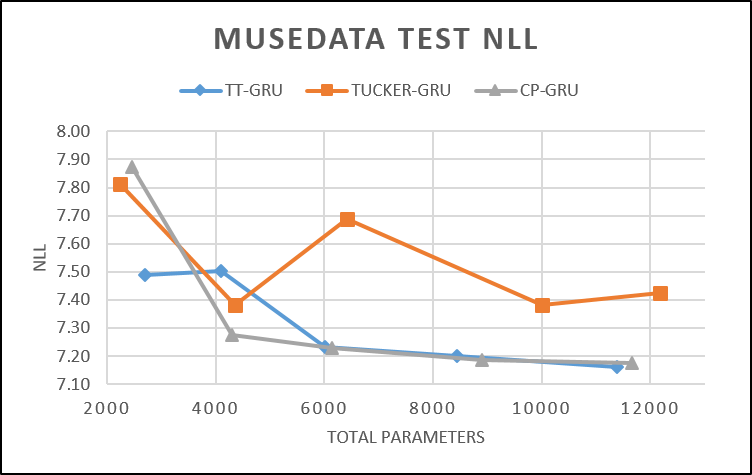}
	\caption{NLL comparison between TT-GRU, Tucker-GRU, and CP-GRU on MuseData test set} 
	\label{fig:plot_musedata}
\end{figure}

We report results of our experiments in Table.~\ref{tbl:allexp}. For the baseline model, we choose standard GRU-RNN without any compression on the weight matrices. For the comparison between compressed models (CP-GRU, Tucker-GRU and TT-GRU), we run each model with 5 different configurations and varied the number of parameters ranged from 2232 up to 12184. In Figure~\ref{fig:plot_nottingham}-\ref{fig:plot_musedata}, we plot the negative log-likelihood (NLL) score corresponding to the number of parameters for each model. From our results, we observe that TT-GRU performed better than Tucker-GRU in every experiments with similar number of parameters. In some datasets (e.g., PianoMidi, MuseData, Nottingham), CP-GRU has better results compared to Tucker-GRU and achieves similar performance (albeit slightly worse) as TT-GRU when the number of parameters are greater than 6000.

\section{Related Work}
Compressing neural network has been studied intensively in the recent years. Some works have been proposed to reduce the number of bits needed to represent neural network weight values. Instead of using full precision 32-bit floating points, Courbariaux et al. \cite{courbariaux2014training} and Gupta et al. \cite{gupta2015deep} half precision floating points is sufficient to represent the neural network weights. Later, Courbariaux et al. \cite{courbariaux2016binarized} represented the floating point numbers in the weight matrices into the binary values and replace most arithmetic operations with bit-wise operations.

``Distilling'' the knowledge from a larger model into a smaller model is popularized by Hinton et al. \cite{hinton2015distilling}. There are several steps for knowledge distillation: 1) Train a large neural network model with hard labels as the output target, 2) Using a trained large neural network, generate the soft label from each input by taking the last softmax output with higher temperature, 3) Train a smaller neural network with the soft target as the output target. Tang et al. \cite{tang2016recurrent} adapt knowledge distillation by using large DNN soft-targets to assist the RNN model training. Kim et al. \cite{kim2016sequence} proposed sequence-level knowledge distillation for compressing neural machine translation models.

Low-rank approximation for representing the weight parameters in neural network has been studied by \cite{sainath2013low}, \cite{denil2013predicting}, \cite{denton2014exploiting}. The benefits from low-rank approximation are reducing the number of parameters as well as the running time during the training and inference stage. Novikov et al. \cite{novikov2015tensorizing} replaced the weight matrix in the convolutional neural network (CNN) final layer with Tensor-Train\cite{oseledets2011tt}(TT) format. Tjandra et al. \cite{tjandra2017compressing} and Yang et al. \cite{yang2017tensor} utilized the TT-format to represent the RNN weight matrices. Based on the empirical results, TT-format are able to reduce the number of parameters significantly and retain the model performance at the same time. Recent work from \cite{ye2017learning} used block decompositions to represent the RNN weight matrices.

Besides the tensor train, there are several tensor decomposition methods that are also popular such as CP and Tucker decomposition. However, both the CP and the Tucker decomposition have not yet been explored for compressing the RNN model. In this paper, we utilized the CP and the Tucker decomposition to compress RNN weight matrices. We also compared the performances between the CP, Tucker and TT format by varying the number of parameters at the same task.

\section{Conclusion}
In this work, we presented some alternatives for compressing RNN parameters with tensor decomposition methods. Specifically, we utilized CP-decomposition and Tucker decomposition to represent the weight matrices. For the experiment, we run our experiment on polyphonic music dataset with uncompressed GRU model and three tensor-based RNN models (CP-GRU, Tucker-GRU and TT-GRU). We compare the performance of between all tensor-based RNNs under various number of parameters. Based on our experiment results, we conclude that TT-GRU has better performances compared to other methods under the same number of parameters.
% use section* for acknowledgment
\section*{Acknowledgment}
Part of this work was supported by JSPS KAKENHI Grant Numbers JP17H06101 and JP17K00237. The authors would like to thank Shiori Yamaguchi for helpful discussions and comments.

% trigger a \newpage just before the given reference
% number - used to balance the columns on the last page
% adjust value as needed - may need to be readjusted if
% the document is modified later
%\IEEEtriggeratref{8}
% The "triggered" command can be changed if desired:
%\IEEEtriggercmd{\enlargethispage{-5in}}

% references section

% can use a bibliography generated by BibTeX as a .bbl file
% BibTeX documentation can be easily obtained at:
% http://mirror.ctan.org/biblio/bibtex/contrib/doc/
% The IEEEtran BibTeX style support page is at:
% http://www.michaelshell.org/tex/ieeetran/bibtex/
%\bibliographystyle{IEEEtran}
% argument is your BibTeX string definitions and bibliography database(s)
%\bibliography{IEEEabrv,../bib/paper}
%
% <OR> manually copy in the resultant .bbl file
% set second argument of \begin to the number of references
% (used to reserve space for the reference number labels box)
\bibliographystyle{IEEEtran}
\bibliography{refs}

% that's all folks
\end{document}